# CASCADED NEURAL NETWORKS WITH SELECTIVE CLASSIFIERS AND ITS EVALUATION USING LUNG X-RAY CT IMAGES

*Masaharu Sakamoto, Hiroki Nakano*

IBM Tokyo Laboratory, Tokyo, Japan

## ABSTRACT

Lung nodule detection is a class imbalanced problem because nodules are found with much lower frequency than non-nodules. In the class imbalanced problem, conventional classifiers tend to be overwhelmed by the majority class and ignore the minority class. We therefore propose cascaded convolutional neural networks to cope with the class imbalanced problem. In the proposed approach, cascaded convolutional neural networks that perform as selective classifiers filter out obvious non-nodules. Successively, a convolutional neural network trained with a balanced data set calculates nodule probabilities. The proposed method achieved the detection sensitivity of 85.3% and 90.7% at 1 and 4 false positives per scan in FROC curve, respectively.

## 1. INTRODUCTION

Lung cancer occupies a high percentage in the mortality rates of cancer even on a worldwide basis [1]. Early detection is one of the most promising strategies to reduce lung cancer mortality [2]. In recent years, along with performance improvements of CT equipment, increasingly large numbers of tomographic images have come to be taken (e.g., at slice intervals of 1 mm), resulting in improvements in the ability of radiologists to distinguish nodules. However, there is a limitation to interpreting a large number of images (e.g., 300–500 slices / scan) by relying on humans. Computer-aided diagnosis (CAD) systems show promise for the urgent task of time-efficient interpretation of CT scans.

Lung nodule classification is a class imbalanced problem because nodules are found with much lower frequency than non-nodules. In the class imbalanced problem, conventional classifiers tend to be overwhelmed by the majority class and ignore the minority class.

As one method to cope with the class imbalanced problem in lung nodule detection, we propose cascaded neural networks with selective classifiers. Our method can achieve a few false positives while maintaining high sensitivity in the lung nodule detection. As a result, it helps decrease the burden of image interpretation on radiologists.

In the proposed approach, concatenated convolutional neural networks that perform as selective classifiers for filtering out obvious non-nodules are followed by a convolutional neural network (CNN) trained with a balanced data set for calculating nodule probabilities.

In this paper, we present the implementation of a selective classifier and demonstrate how the cascaded neural networks perform as selective classifiers and contribute to reducing the false positives while maintaining high sensitivity. We then compare the performance with a conventional CNN approach.

## 2. PRIOR WORK

### 2.1. Computer-aided diagnosis

Computer-aided diagnosis (CAD) systems are crucial for time-efficient interpretation of CT scans. In one study [2], six computer-aided diagnosis algorithms of lung nodules in computed tomography scans were compared. These methods extract features in lung nodule images with a signal processing technique and classify nodule candidates by using pattern matching based on statistics or a machine learning method such as the k-nearest neighbor algorithm (k-NN) and neural networks.

In recent years, spurred by the large amounts of available data and computational power, CNNs have outperformed state-of-the-art techniques in several computer vision applications [3]. This is because CNN can be trained end-to-end in a supervised fashion while learning highly discriminative features, thus removing the need for handcrafting nodule descriptors.

Setio et al. [4] used a CNN specifically trained for lung nodule detection. On 888 scans of a publicly available data set, their method reached high detection sensitivities of 85.4% and 90.1% at 1 and 4 false positives per scan in FROC curve, respectively.

### 2.2. Class imbalanced problem

Lung nodule detection is a class imbalanced problem, as nodules are found with much lower frequency than non-nodules. In other words, many irregular lesions that are visible in CT images are non-nodules, such as blood vessels or ribs. In the class imbalanced problem, conventional classifiers tend to be overwhelmed by the majority class and ignore the minority class.

Japkowicz [6] showed that oversampling the minority class and subsampling the majority class are both very effective methods of coping with the problem.

## 3. CASCADED NEURAL NETWORKS

The aim of this study is to develop a lung nodule classifier with a few false positives while maintaining high sensitivity in order to decrease the burden of image interpretation on radiologists. To achieve this aim, we propose cascaded neural networks consisting of CNNs that perform as selective classifiers for filtering out obvious non-nodules such as blood vessels or ribs followed by a CNN trained with a balanced data set for calculating nodule probabilities.

Figure 1 shows the 12-layer CNN we utilized. This same CNN was used in all experiments in this study. Figure 2 shows a schematic diagram of the cascaded neural networks we propose. The CNNs are concatenated in cascade arrangement. $S_1$, $S_2$, ... $S_n$ are CNNs that perform as selective classifiers, classifying non-nodules into obvious non-nodules and suspicious nodules. To implement such selective classifiers, the CNNs are trained with an inversed imbalanced data set consisting of many nodule images and a few non-nodule images. By "inverse" we mean that the ratio of the number of nodules and non-nodules is reversed against the original data set.

At $S_1$, by using the CNN that performs as selective classifier, the data set is classified, and then, the nodules whose probabilities fall below a threshold are removed from the data set. The threshold value is determined from a standard deviation σ of the nodule probability distribution of the data set. 0.25σ is used in this study.

At $S_2$, the same procedure described above is applied again to remove the obvious non-nodules from the data set. By cascading the selective classifiers further, the number of non-nodules decreases to balance with the number of nodules.

In the last stage $S_{n+1}$, the CNN is trained by a balanced data set extracted from the data set at $S_n$. By "balanced" we mean that the number of nodules is almost equal to the number of non-nodules. Finally, the CNN calculates the probabilities of the nodules in the data set.

The unique point of our method, as described above, is that it uses cascaded CNNs that perform as selective classifiers. Viola-Jones [7] and Jianxin et al. [8], in contrast, use weak classifiers.

## 4. EXPERIMENTS

### 4.1. Lung CT image data set

We use the lung CT scan data set obtained from Lung Nodule Analysis 2016 [5]. This set includes 888 CT scan images along with annotations that were collected during a two-phase annotation process overseen by four experienced radiologists. Each radiologist marked lesions they identified as non-nodule, nodule < 3 mm, and nodule >= 3 mm. The data set consists of all nodules >= 3 mm accepted by at least 3 out of 4 radiologists. The complete data set is divided into ten subsets to be used for the 10-fold cross-validation. For convenience, the corresponding class label (0 for non-nodule and 1 for nodule) for each candidate is provided. 1,348 lesions are labeled as nodules and the other 551,062 are non-nodule lesions. In this study, center coordinates of each lesion are given.

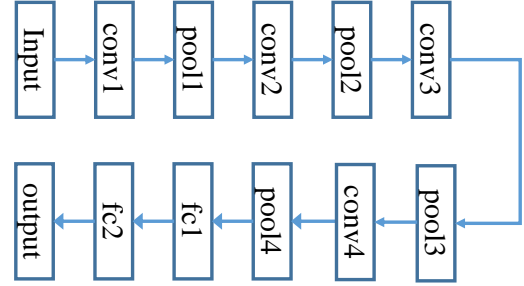

**Figure 1. 12-layer convolutional neural network.**

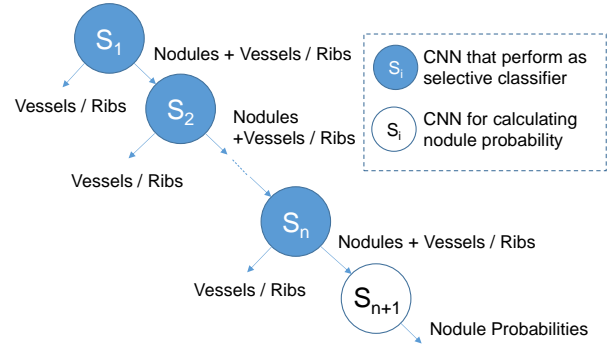

**Figure 2. Schematic diagram of cascaded CNNs. $S_1$, $S_2$, ... $S_n$ are CNNs that perform as selective classifiers to filter out non-nodule lesions. $S_{n+1}$ is a CNN to obtain nodule probabilities. All CNNs have the same structure as shown in Fig. 1.**

We use three consecutive slices to obtain volumetric information. Examples of non-nodule images and nodule images in the data set are given in Figure 3. Each image size cropped from CT scan images is 48 pixels × 48 pixels.

### 4.2. Selective classifier

For full use of the data set, 10-fold cross-validation is used. For training CNNs as selective classifiers, nine subsets are used, and one subset is used for the test. By using the cross-validation, a total of ten classifiers are produced. To make the training data for implementing selective classifiers, non-nodules in a subset are subsampled to 200 samples, and nodules are oversampled nine times by randomly rotating and scaling original images. As a result, the number of

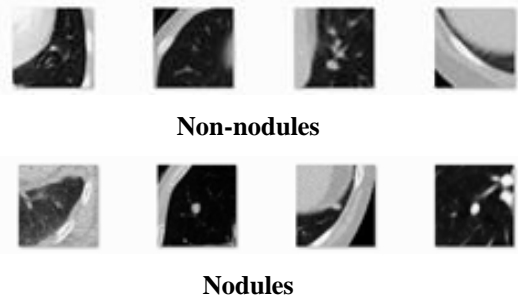

Non-nodules

Nodules

Figure 3. Example of lesion images in data set of Lung Nodule Analysis 2016.

oversampled nodules is about six times the number of subsampled non-nodules. The ratio of the nodules and the non-nodules is inverted from the original data set.

Ten CNNs that perform as selective classifiers are used to evaluate the data set in the ten folds. Figure 4 shows the histogram of the nodule probabilities of the data set. By using the selective classifiers, most of the nodules (class 1) concentrate around probability 1.0, as shown in Fig. 4(a), while the non-nodules (class 0) are separated around probabilities 0.0 and 1.0, as shown in Fig. 4(b). We assume the samples around probability 0.0 are accepted as non-nodules and the samples around probability 1.0 are nodules.

The nodule candidates where the probabilities fall below a threshold are classified as obvious non-nodules and then removed from the data set. The threshold value is determined from a standard deviation $\sigma$ of the non-nodule probability distribution of the data set. In the case shown in Fig. 4(b), the standard deviation is about 0.4 and then the threshold is about 0.1. Several samples with low probabilities in the nodule class (class 1 in Fig. 4(a)) are accidentally removed from the data set as a side effect. This is what causes the false negatives.

### 4.3 Baseline classifier

To obtain a baseline performance of lung nodule detection, ten CNNs are trained using a balanced data set with subsampled non-nodules and oversampled nodules with 10-fold cross-validation. The nodules are oversampled ten times by randomly rotating and scaling original images and non-nodules are subsampled to balance the number of samples with nodules. The structure and parameters of the CNNs are the same as the selective classifiers.

Figure 5 shows the histogram of the nodule probabilities of the data set. Most of the nodules are concentrated around 1.0 in probability, but a smaller concentration is also seen around 0.0, as shown in Fig. 5(a). Most of the non-nodules concentrate around 0.0, but a little concentration is also seen around 1.0, as shown in Fig. 5(b).

### 4.4. Experimental results

Table 1 shows the number of lesions after filtering out non-nodules by each number of cascaded selective classifiers. The

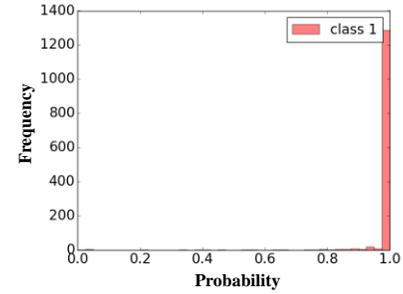

(a)

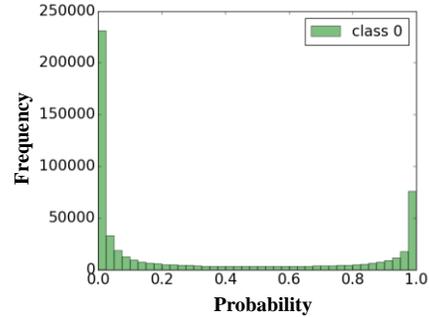

(b)

Figure 4. Histogram of nodule probabilities of nodule candidates by selective classifiers. (a) Nodules and (b) Non-nodules.

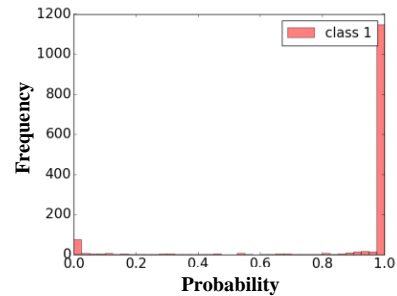

(a)

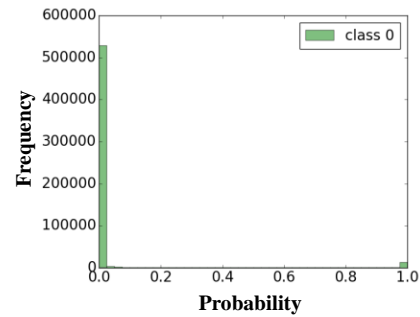

(b)

Figure 5. Histogram of nodule probabilities of nodule candidates by CNNs trained with balanced data set. (a) Nodules and (b) Non-nodules.

**Table 1. Number of lesion images in different number of selective classifiers**

| No. of selective classifiers | 1 | 2 | 3 | 4 | 5 |
|---|---|---|---|---|---|
| No. of non-nodules | 188966 | 76289 | 36728 | 18322 | 11265 |
| No. of nodules | 1344 | 1331 | 1311 | 1302 | 1290 |

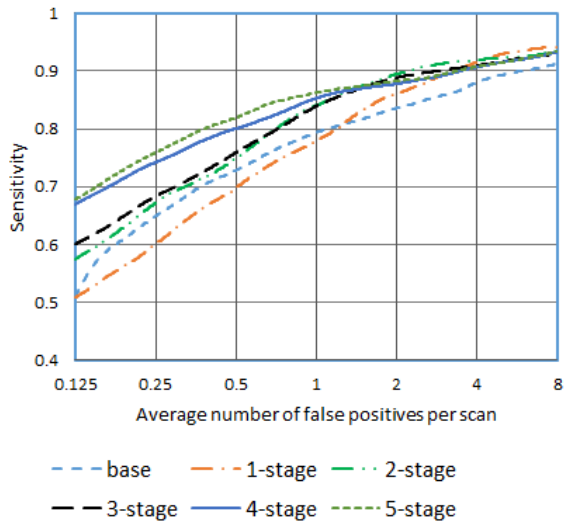

**Figure 6. FROC curves of cascaded CNNs with different number of selective classifiers and a CNN with balanced data set (baseline).**

number of non-nodules is reduced from 551,062 to 188,966 at the first stage. By using five cascaded selective classifiers and obvious non-nodules filtering, the ratio of non-nodules and nodules shrinks to about 10 times at stage 5 from about 400 times of the original data set.

Figure 6 shows FROC curves of different numbers of cascaded selective classifiers and the baseline classifier trained by the balanced data set described in the previous section. The baseline performance achieves the detection sensitivity of 79.3% and 88.1% at 1 and 4 false positives per scan, respectively. The detection sensitivity of just one selective classifier is comparable with the baseline, but by cascading more selective classifiers, the detection sensitivity is improved. The 4-stage selective classifiers reach the detection sensitivity of 85.3% and 90.7% at 1 and 4 false positives per scan, respectively. However, at the same time, dozens of nodules are also filtered out. This is what causes the false negatives. Considering the false negatives, the 4-stage selective classifiers had the best performance among those tested.

## 5. CONCLUSION

In this paper, we have presented cascaded neural networks with selective classifiers to reduce the false positives of lung nodule detection in CT scan images. We have shown that the proposed method achieves good results for the lung nodule detection task in comparison with a conventional CNN approach. By using the cascaded CNNs, obvious non-nodules are rejected at each cascading stage. The system thus presents nodule candidates with nodule probabilities to radiologists, which suggests that the system can decrease the burden of image interpretation on radiologists.

As a future work, we will optimize the hyper-parameters of CNN (batch size, dropout rate, etc.) to reduce false negatives.

## 6. REFERENCES

[1] Cancer Research: Lung cancer mortality statistics. Retrieved Jun 24, 2016, http://www.cancerresearchuk.org/health-professional/cancer-statistics/statistics-by-cancer-type/lung-cancer/mortality.
[2] B.V. Ginneken, "Comparing and combining algorithms for computer-aided detection of pulmonary nodules in computed tomography scans: The ANODE09," study, Medical Image Analysis 14, 707–722 (2010).
[3] A. Krizhevsky, I. Sutskever and G.E. Hinton, "ImageNet classification with deep convolutional neural networks," Advances in Neural Information Processing Systems 25, 1097–1105 (2012).
[4] A.A.A. Setio, F. Ciompi, G. Litjens, P. Gerke, C. Jacobs, S.J. van Riel, M.M.W. Wille, M. Naqibullah, C.I. Sánchez, and B. van Ginneken, "Pulmonary Nodule Detection in CT Images: False Positive Reduction Using Multi-View Convolutional Networks," IEEE Transactions on Medical Imaging, Vol. 35, No. 5, pp. 1160 - 1169 (2016).
[5] Lung Nodule Analysis 2016. Retrieved Jun 24, 2016, http://luna16.grand-challenge.org/
[6] N. Japkowicz, "Learning from Imbalanced Data sets: A Comparison of Various. Strategies," AAAI Technical Report WS-00-05 (2000).
[7] P. Viola and M. Jones, "Rapid object detection using a boosted cascade of simple features," Computer Vision and Pattern Recognition, 2001. CVPR 2001. Proceedings of the 2001 IEEE Computer Society Conference on, Vol. 1, pp. 511–518 (2001).
[8] J. Wu, J.M. Rehg and M.D. Mullin, "Learning a rare event detection cascade by direct feature selection," Proceedings of Advances in Neural Information Processing Systems (NIPS) 16, pp. 1523-1530 (2004).